%% file: root.tex
\documentclass[letterpaper, 10 pt, conference]{ieeeconf}  

\IEEEoverridecommandlockouts                              

\usepackage{amsmath} 
\usepackage{amssymb}  
\usepackage{graphicx} 
\usepackage{multirow}
\usepackage{verbatim}

\title{\LARGE \bf
Joint torques prediction of a robotic arm using neural networks
}

\author{Giulia d'Addato$^{1,5}$, Ruggero Carli$^{2}$, Eurico Pedrosa$^{3}$, Artur Pereira$^{3}$, Luigi Palopoli$^{4}$, Daniele Fontanelli$^{5}$ 
\thanks{}
\thanks{$^{1}$ Project co-funded by the European Union – Next Generation Eu - under the National Recovery and Resilience Plan (NRRP), Mission 4 Component 1 Investment 4.1 - Decree No. 118 (2/03/2023)  of Italian Ministry of University and Research - Concession Decree No. 2333 (22/12/2023) of the Italian Ministry of University and Research, Project code D93C23000470005, within the Italian National Program PhD Programme in Autonomous Systems (DAuSy).}%
\thanks{$^{2}$ Department of Information Engineering, Universit\`a di Padova, Italy.}%
\thanks{$^{3}$ IEETA, DETI, LASI, University of Aveiro, Portugal.}
\thanks{$^{4}$ Department of Information Engineering and Computer
  Science, Universit\`a  di Trento, Italy.}
\thanks{$^{5}$ Department of Industrial Engineering, Universit\`a di
  Trento, Italy.} %
}

\begin{document}

\maketitle
\thispagestyle{empty}
\pagestyle{empty}

\begin{abstract}
  Accurate dynamic models are crucial for many robotic applications. Traditional approaches to deriving these models are based on the application of Lagrangian or Newtonian mechanics. Although these methods provide a good insight into the physical behaviour of the system, they rely on the exact knowledge of parameters such as inertia, friction and joint flexibility. In addition, the system is often affected by uncertain and nonlinear effects, such as saturation and dead zones, which can be difficult to model.  A popular alternative is the application
  of Machine Learning (ML) techniques - e.g., Neural Networks (NNs) - in the context of a ``black-box" methodology.
  This paper reports on our experience with this approach for a real-life 6 degrees of freedom (DoF) manipulator. Specifically, we considered several NN architectures: single NN, multiple NNs, and cascade NN. We compared the performance of the system by using different policies for selecting the NN hyperparameters. 
  Our experiments reveal that the best accuracy and performance are obtained by a cascade NN, in which we encode our prior physical knowledge about the dependencies between joints, complemented by an appropriate optimisation of the hyperparameters. 
  
\end{abstract}


\input{Introduction}

\input{Problem}
\input{Experiments}
\input{Conclusions}

\bibliographystyle{IEEEtran}
\bibliography{root}

\end{document}

%% file: Introduction.tex
\section{INTRODUCTION}

One of the major problems in the robotics field is the identification of dynamic models.
The problem is becoming increasingly important in a class of emerging applications where the control of contact forces and torques between the robot and the environment is a crucial safety concern~\cite{close2001modeling, vicentini2021collaborative}.
For instance, in the context of human-robot cooperation, safety is essential and requires the implementation of appropriate algorithms to control and limit the forces exerted on the operator when s/he comes into contact with the robot~\cite{inproceedings, 7942795}. 
An accurate knowledge of the torques of each joint is key to the design of this type of controller and is the main motivation behind the present work. 

The most immediate way of reconstructing the information about the joint torques is by direct measurement. This solution requires a set of sensors placed at each joint.  
However, this approach is expensive, firstly because the hardware is expensive in its own right, and secondly because the system requires frequent maintenance operations to recalibrate the sensors.

A possible alternative is to model the dynamics of the system, which relates the torques to the joint state variables (see Equation \eqref{standardEq}).
This approach is affected by two main issues: (1) the physical parameters are difficult to identify due to the nonlinearity of the system; (2) there are several sources of uncertain dynamic effects (e.g., dead zones and saturation), which are unknown or difficult to model.

In these cases, ML techniques come to the rescue. Specifically, modelling the system by using a black-box approach allows us to use an NN to learn the complex relationship within the data, also generalising across a wide range of scenarios and adapting to changes in the system configuration. 
In particular, NNs hold the promise to capture the workings of factors such as friction, inertia and nonlinear behaviour. 
This promising avenue requires a proper understanding of
the most suitable NN architecture for the problem at hand and of its hyperparameters.

In this paper, we investigate the aforementioned problem in a specific application scenario: the prediction of the joint torques of a 6 DoF industrial robotic arm.
Our key contribution lies in the design and implementation of three distinct NN architectures. 
The core problem is how to exploit the dependencies between the joints, and each of the three solutions addresses this in a different way. 
The first architecture neglects these dependencies altogether and attempts to learn them from the data. 
The second one takes the dependencies into account by employing three separate NNs for each joint group.
The third one adopts a hybrid approach that maintains joint groupings while also incorporating inter-group dependencies. 

The selection of the optimal model is influenced by a cost function that evaluates the prediction error and depends on the choice of NN hyperparameters.
In this work, we considered  the prediction error as a performance metric and our experimental results show that the hybrid architecture is the most accurate one when accompanied by an adequate optimisation of the hyperparameters.

%

%

\section{Related works}

The development of human-robot collaboration is leading to high safety
requirements.  Since the need for interaction and proximity to the
user requires the robotic arm to be aware of its surroundings, motion
control and force computation must be as accurate as possible,
especially when the robot is moving at high speed. The problem of
force/torque prediction in modern robotic assistants has been
approached from various perspectives. In general, the difficulty of
the prediction is due to the nonlinear dynamics, the uncertainties in
the parameter values and the disturbances not considered in the
analytical model of the robot dynamics. In order to overcome these
issues, several approaches have been proposed in the literature,
including observer-based, least-squares and learning-based
methods~\cite{myrvold2020estimating}.

In most modern applications, external force sensors are mounted on the
robot to estimate forces. However, these are expensive tools and not
all robots are equipped with them. Therefore, cheaper sensor-less
estimation methods are often used. A commonly used technique for
predicting external torques consists of force observers or generalised
momentum observers, which isolate the external torques on the joint
torques, without needing external force
sensors~\cite{gold2019external}. The prediction of the forces acting
on the manipulator is based on the measured position and orientation
data and on the knowledge of the control forces and moments,
remodelling the dynamics of the observer error from time to time.

This method can be extended and combined with learning-based
approaches, as ML is a growing field of research that is also showing
promising results in torque prediction. The advantage of this type of
methods is not only that they do not require the knowledge of an
accurate analytical model (as they treat the function approximation
task as a black-box), but also that they are computationally
cheaper. There are several standard ML techniques such as Linear
Regression, Gaussian Mixture Regression, Gaussian Process Regression,
Support Vector Regression, Feedforward or Recurrent NNs: all these
models learn a forward or inverse mapping from the joint states to the
joint torques~\cite{shalev2014understanding, zakeri2018estimating}. An
interesting class of black-box solutions is based on Gaussian process
regression (GPR), where each joint is usually considered individually
and the unknown function, i.e. its torque, is modelled a priori as a
single Gaussian process (GP)~\cite{Libera_2020,
  giacomuzzo2022advantages}.

A promising novel approach that uses ML or Deep Learning (DL)
counteracts the loss of physical plausibility of black-box
representations by embedding physical constraints in the network
design. For instance, a recently implemented solution to this issue is
the so-called Deep Lagrangian Network (DeLaN)~\cite{lutter2019deep},
which integrates Lagrangian mechanics into DL frameworks, thus
combining the solid theoretical control and physics foundations with
the flexibility of DL. By imposing Lagrangian mechanics and energy
conservation principles on deep networks, a DeLaN can learn physical
attributes such as the mass matrix, centripetal, Coriolis,
gravitational, and frictional forces, and system energy. A similar but
more general technique for learning arbitrary Lagrangians is the
so-called Lagrangian neural network (LNN)~\cite{cranmer2020lagrangian}. These approaches ensure that the learned parameters align more closely with real-world physics, enhancing the reliability and interpretability of the model's outputs. 

Unlike the latter approaches, which can be critical in several applications such as control and simulation, our approach is based on feedforward neural models that approximate the relationship between the joint states and the joint torques. This guarantees a robust and reliable solution, particularly suitable in contexts where the experimentation is limited or the robustness of the approach is prioritised. Furthermore, while techniques such as DeLaN and LNN are relatively new and less tested, the approach we adopt is well documented in the literature, confirming its reliability and robustness.

%% file: Problem.tex
\section{PROBLEM FORMULATION AND METHODS}

Finding a dynamic model that accurately estimates the joint torques
$\boldsymbol{\tau}$ of a robotic arm is a crucial and complex task.  Robot dynamics
consists of the relationship between the forces acting on a robot and
its resulting motion. This relation can be written as an
inverse dynamic equation, where the torques being applied to the
joints are defined as a function of the joints state, namely their
position, velocity and acceleration ($\mathbf{q}, \dot{\mathbf{q}}, \ddot{\mathbf{q}}$).
A classical way to draw up this inverse dynamics equation is the
Euler–Lagrange formulation, which leads to the following standard
equation:
\begin{equation}\label{standardEq}
\tau = B(\textbf{q}) \ddot{\textbf{q}}+C(\textbf{q}, \dot{\textbf{q}})+g(\textbf{q})+h(\textbf{q}, \dot{\textbf{q}}, \ddot{\textbf{q}})  ,
\end{equation} 
where $B(\textbf{q}) \ddot{\textbf{q}}$ represents the inertial terms,
$C (\textbf{q}, \dot{\textbf{q}}) \dot{\textbf{q}}$ the centripetal
and Coriolis terms, $g(\textbf{q})$ the gravity terms,
$h(\textbf{\textbf{q}}, \dot{\textbf{q}}, \ddot{\textbf{q}})$ any
torques related to other additional joint dynamic effects, such as
friction, and $\tau$ the effect of any force exerted by the external
environment on the robot.

\begin{figure}[t]
\centering
    \includegraphics[width=0.48\textwidth]{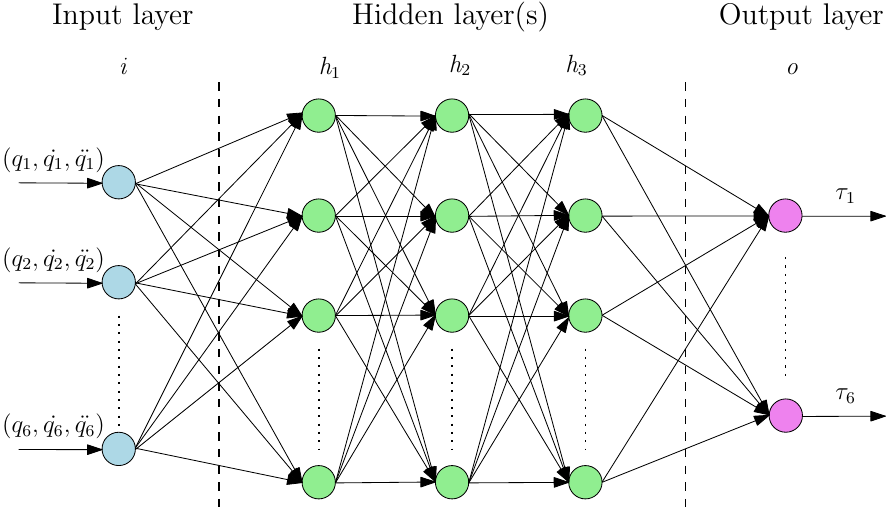}
\caption{\label{fig:NNarch} Example of feedforward neural network architecture}
\end{figure}

The analytical estimation of the dynamic model often requires
simplifications in order to reduce the computational
complexity. Indeed, inertia and friction parameters can be difficult
to obtain analytically in an accurate way. For this reason, ML
approaches can be helpful and effective to identify the dynamics. In
this work, no analytical dynamics derivation is presented, but the
procedure would require to calculate quantities such as Lagrangian,
potential and kinetic energy and to solve the standard
dynamics~\eqref{standardEq}, which can be more compactly expressed as:
\begin{equation}
  \label{invDyn}
  \boldsymbol{\tau} = f(\mathbf{q}, \dot{\mathbf{q}}, \ddot{\mathbf{q}}). 
\end{equation}
The function $f(\cdot)$ is the nonlinear relationship to be found by
training an NN (as in Figure~\ref{fig:NNarch}) on the given dataset.
This data is generated by moving the actual robot in space and
collecting data for positions, velocities, accelerations and torques
from each joint.  While $\mathbf{q}$, $\dot{\mathbf{q}}$ and
$\ddot{\mathbf{q}}$ are used for the NN training inputs,
$\boldsymbol{\tau}$ represents the outputs, consistent with the
dynamics~\eqref{invDyn}. Since the data are collected from the actual
mechanism, also frictional terms are automatically accounted for by
the NN, thus solving one of the most relevant issues of traditional
nonlinear identification approaches.  By comparing the NN torque
estimation against the actual values, the trained network accuracy can
be evaluated based on the prediction reliability. For this purpose,
one of the most common performance metrics, also used in this work, is
the Mean Squared Error (MSE), which is defined as:
\begin{equation}
  MSE = \frac{1}{n} \sum_{i=1}^n\left(\boldsymbol{\tau}_i-\hat f\left(\mathbf{q}_i, \dot{\mathbf{q}}_i, \ddot{\mathbf{q}}_i \right)\right)^2 ,
\end{equation}
where $\hat f(\cdot)$ is the learned model generating the
prediction $\hat{\boldsymbol{\tau}}_i$~\cite{geron2022hands}.

\subsection{Data acquisition and pre-processing} \label{dataAcquisition}

The first step toward the construction of our NNs is to acquire and
pre-process the dataset. The data collection process is carried out
following the grid search principles for each of the six joints, so as
to optimise both the data size and the acquisition time.  This
optimisation step allows us to manage large amount of data. Redundant
data are removed, by grouping joints into independent subgroups based
on orthogonal motions. In the specific case of the arm used for this
work, the arm's joints are categorised into three groups: (a) joint 1;
(b) joints 2 and 3; (c) joints 4, 5 and 6. This grouping aligns with
the structure of the NN models proposed in the next section.
Data collection involves setting initial, final, and step angles for
each joint, moving one joint at a time within each subgroup. To
accelerate the acquisition and to avoid excessive motions, the joints are not
reset to their initial position, but they are moved sequentially. The
first joint does not affect any of the other ones, as it is
fixed at the base and its configuration is generally not changed
during the data acquisition phase, except when the speed is
changed. Finally, shuffling the configurations ensures that a large
number of poses is used in the learning phase.

The dataset includes joint position, velocity, acceleration, and torque for each of the 6 joints at every instant of acquisition. Before entering the training phase, we refined the dataset by applying a a standardisation procedure.
Standardisation amounts to normalise the variable ranges, by transforming the values $X$ of a dataset so that the mean of the training samples is 0 ($\mu = 0$) and the standard deviation is equal to one ($\sigma = 1$):
\begin{equation}
X_{new} = \frac{X - \mu}{\sigma} .
\end{equation}  
This refinement is known to enhance dataset manageability and algorithmic efficiency~\cite{misra2019impact}. 
In this work, data normalisation is performed using the \texttt{StandardScaler()} function from the sklearn library in Python.

\subsection{Neural Networks models}

NNs are able to capture both the nonlinear dependencies and the
real-world effects (e.g., friction, dead bands, etc.) that are not
easy to incorporate in the analytical dynamics (Equation~\eqref{standardEq}).
In this work, the measurements in the dataset are acquired from a 6
DoF robotic arm, so the network could potentially have 18 inputs,
i.e. position, velocity and acceleration for each joint, and 6
predicted outputs. In fact, we can exploit our prior knowledge of the
dependencies between data and their significance in the overall model,
in order to both reduce the effort of the neural implementation, and
to improve the accuracy. The implementation of three distinct neural models is indeed aimed at understanding the impact of joint
dependencies on torque prediction. These architectures include a
network that completely ignores joint dependencies, another that fully
accounts for dependencies by separating them into three distinct
networks, and a hybrid model that preserves joint groupings while
keeping them connected. Through this approach, we seek to understand
the extent to which joint dependencies contribute to improve torque
prediction accuracy.
The three NN implementations tested include:

\begin{enumerate}

\item Single fully connected neural network: it receives 17 input
  parameters, i.e. all the available joint variables except the
  position of the first joint. The latter is removed from the input
  data as it is independent of the outputs and therefore does not
  provide significant information for the network learning process. Its structure is similar to that in Figure~\ref{fig:NNarch};

\item Multiple neural networks corresponding to the independent groups
  of joints: the input variables corresponding to each of the three
  independent groups do not affect the output of the others. This
  model reduces the computational cost as fewer weights need to be
  calculated and updated;

\item Cascade neural network: it starts learning with one input node
  and then adds new inputs as new hidden neurons evolve
  (see Figure~\ref{cascade}).
  \begin{figure}[t]
    \centering \includegraphics[width=0.49\textwidth]{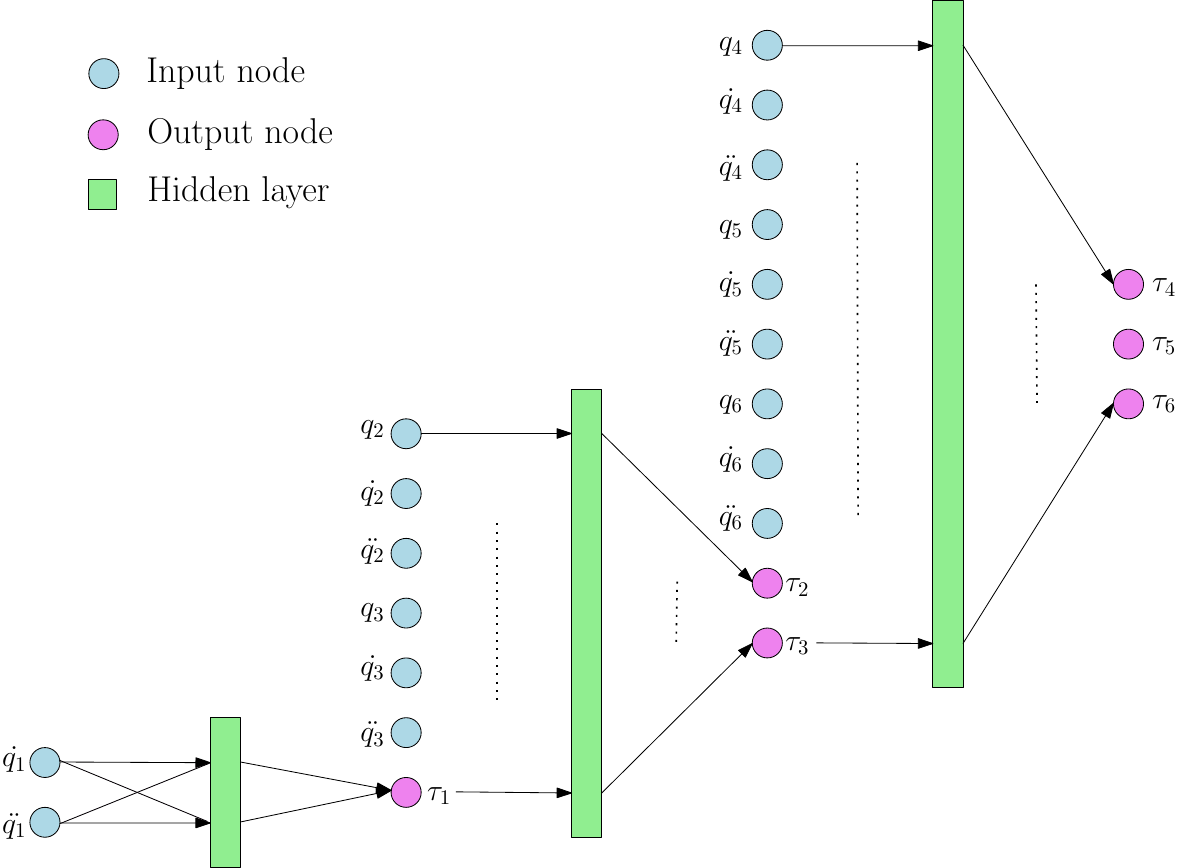}
    \caption{Cascade NN architecture}
    \label{cascade}
  \end{figure}
  This model considers the same groups of joints of the second one,
  but starts by giving as input only the state of the first joint,
  then the output obtained from the latter and the second group of
  joint states, and so on. In practice, the output of each network
  becomes the input of the next one, creating a series of connected
  networks.
\end{enumerate}
Regardless of the architecture chosen, all the networks implemented are multilayer perceptron NNs, connecting multiple layers in a directed graph and using backpropagation as a supervised learning technique~\cite{popescu2009multilayer}.

\subsection{Hyperparameters optimisation} \label{optimOptuna}

Despite rapid advances in ML and DL, practical aspects such as hyperparameter tuning remain challenging~\cite{hyperpar}. Hyperparameters, which define the architecture of the neural model, must be specified prior to training and have a significant impact on model performance.
To achieve the optimal model, hyperparameters must be carefully optimised, a complex task due to their large number. The challenge is to find a solution that efficiently optimises each hyperparameter in the shortest time and consuming the least amount of resources possible.

Key NN hyperparameters include the number of hidden layers and the number of nodes per layer, which determine the complexity and the performance of the network. In addition, the learning rate, which controls the model's response to estimated errors during weight updates, is a critical parameter. Finding the right balance between small and large learning rates affects training efficiency.
Other tunable hyperparameters include dropout rates, optimisation algorithms such as Adaptive Moment estimation (Adam), Stochastic Gradient Descent (SGD), Root Mean Squared Propagation (RMSProp), and so on. For this work, we use PyTorch, an open source framework for ML, and we exploit its tensor class for GPU-accelerated multidimensional array operations. It also introduces several libraries to support the development of Artificial Intelligence (AI) applications, including Optuna, which automates the search for hyperparameters and is exploited in this work~\cite{akiba2019optuna}.

%% file: Experiments.tex
\section{EXPERIMENTS AND RESULTS}

\begin{figure}[t]
\centering
    \includegraphics[width=0.48\textwidth]{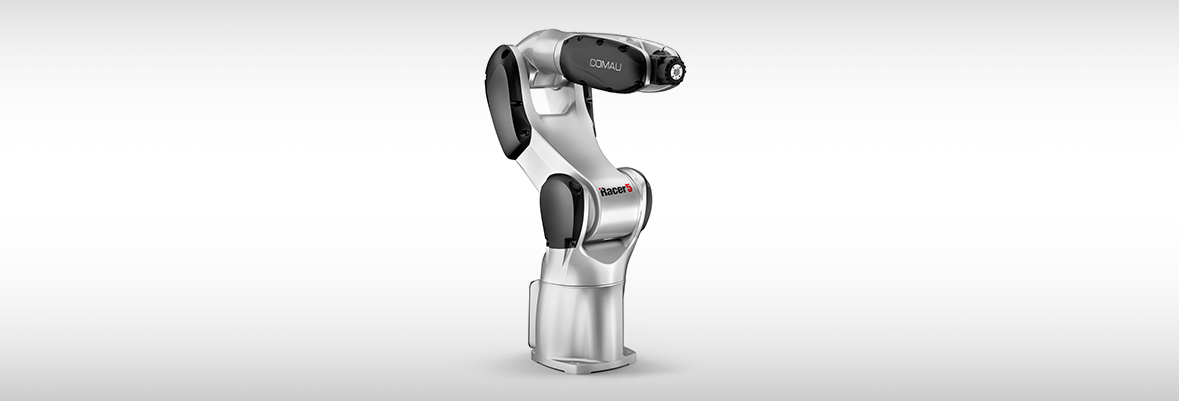}
\caption[]{\label{fig:comau} Comau Racer-5-0.80 robot~\cite{WinNT}}
\end{figure}

The following sections report and analyse the results of the
experiments carried out on the acquired data.  The data used in the
developed models were obtained from the real manipulator shown in
Figure~\ref{fig:comau}, the Comau Racer-5-0.80~\cite{WinNT}, an
anthropomorphic robot with $6$~DoF, guaranteeing fluidity of movement,
high performance in terms of accuracy and repeatability, and safety.

\subsection{Training results}

As a first experiment, the training parameters and network
hyperparameters have been tested and chosen by trial and error and are
listed in Table~\ref{tab:chosenHyperpar}. All tested networks have one
hidden layer and have been trained by removing the position of the
first joint from the input dataset, as explained in
Section~\ref{dataAcquisition}.
Table~\ref{tab:testLoss} shows the results obtained by testing the three different types of network, both with and without pre-processing of the training dataset and trained in 10 epochs. The results consist of the test loss expressed in terms of MSE.
In particular, the average of the total test loss in the 10 epochs is reported here.
\begin{figure*}[t]
\centering
\includegraphics[width=1\textwidth, keepaspectratio]{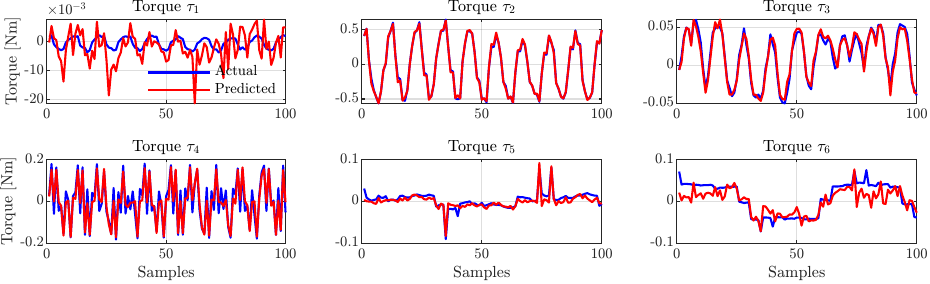}
\caption[Actual and predicted torque values using a single NN after scaling ]{\label{fig:actpredscaling} Actual and predicted torque values using a single NN after scaling}
\end{figure*}
\begin{table}[t] 
\caption{\label{tab:chosenHyperpar}Chosen neural network hyperparameters}
\begin{center}
    \begin{tabular}{| c || c |}
    \hline 
        \multirow{2}{*}{Number of hidden nodes}   & 30 (single NN, cascade NN)\\   & 5,15,30 (Multiple NNs)\\ 
        \hline
        Activation function   & Leaky ReLU \\
        \hline
        Optimisation algorithm  & Adam\\
        \hline
        Learning rate   & 0.001\\ 
        \hline
        Loss function  & MSE\\
        \hline
        Train Test split  & 70:30\\ 
        \hline
        Number of epochs   & 10 \\
        \hline
    \end{tabular}
    
\end{center}

\end{table}

\begin{table}[t]
\caption{Average test MSE in 10 epochs with and without scaling}
\begin{center}
\begin{tabular}{| c || c | c |}
\hline
& Scaling & Average test MSE \\ \hline \hline
\multirow{2}{*}{Single NN}  & without & $2.9953\!\cdot\!10^{-4}$                     \\  \cline{2-3}   
 & with & $2.4363\!\cdot\!10^{-4}$   \\ \hline
\multirow{2}{*}{Multiple NNs}  & without &  $3.4969\!\cdot\!10^{-4}$                        \\ \cline{2-3}
 & with & $2.4445\!\cdot\!10^{-4}$   \\ \hline
 \multirow{2}{*}{Cascade NN}  & without & $2.9951\!\cdot\!10^{-4}$                        \\ \cline{2-3}
 & with & $2.5096\!\cdot\!10^{-4}$  \\ \hline

\end{tabular}
\label{tab:testLoss}
\end{center}
\end{table}

\begin{figure}[t]
\centering
    \includegraphics[width=\columnwidth, keepaspectratio]{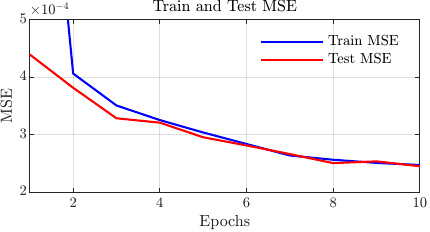}
\caption[Train and test average MSE using a single NN after scaling]{\label{fig:err} Train and test average MSE using a single NN after scaling}
\label{fig:msezoom}
\end{figure}

Figure~\ref{fig:msezoom} shows for instance the evolution of the average MSE after training a single NN.
In general, the graph shows a good fit, since as the epochs progress, both training and test losses decrease, until they become very similar and stabilise around the same value, once the model has sufficiently learned to generalise.
Figure~\ref{fig:actpredscaling} instead compares the actual torques with those predicted by using the first implemented neural model.
Clearly, predicting the first and the sixth torques is quite critical:
while the reason why the former is more difficult to predict has been explained
above, for the latter, being a free joint with no load on it, the data
are not reliable and thus its torque computation is crucial.  The same
model is therefore tested also by removing the state variables of the
sixth joint.  As expected, its removal not only reduces the
complexity, but also decreases the full loss, reaching an MSE value of
$2.2470\!\cdot\!10^{-4}$.
In any case, the consistent accuracy of predictions achieved through a single NN suggests that the combined effects of position, velocity and acceleration values may significantly influence the overall torque outcomes.
%
Moreover, in general, the results obtained after standardising the input data show the usefulness of pre-processing for large datasets, as the loss is slightly lower and the convergence is faster.

\subsection{Optimised networks}

Network improvement can be achieved through an optimisation process (see Section~\ref{optimOptuna}). For the purpose of this project, Optuna, an open source hyperparameter optimisation framework, is used to automate the hyperparameter search by trial and error and to achieve higher performance. In particular, it uses a Bayesian optimisation algorithm called Tree-structured Parzen Estimator (TPE), which is the default sampler in Optuna and uses the history of previously evaluated hyperparameter values to sample the next ones.
This sampler, based on independent sampling, fits a Gaussian Mixture Model (GMM), for each trial and for each parameter, to the set of parameter configurations associated with the best possible objective value, in this case corresponding to the test loss.

In this work, the following three hyperparameters were optimised:

\begin{enumerate}
    \item Number of hidden nodes: from 10 to 50;
    \item Optimisers: SGD, Adam, RMSprop;
    \item Learning rates: from $1\!\cdot\!10^{-4}$ to $1\!\cdot\!10^{-1}$.
\end{enumerate}
The objective function to be optimised, minimised in our case, is the total test loss in terms of MSE.
The process involves 10 trials and produces the results shown in Table~\ref{tab:optimHyper} for each neural structure.
\begin{table}[t]
\caption{\label{tab:optimHyper}Optimised hyperparameters}
\begin{center}
    \begin{tabular}{| c || c | c | c |}
    \hline
        \multirow{2}{*}{NN architecture} & Number of & Optimisation & \multirow{2}{*}{Learning rate} \\ & hidden nodes & algorithm & \\

        \hline \hline
        Single NN & 36 & Adam & $1.795849\!\cdot\!10^{-3}$\\
        \hline
        Separate NNs & 49, 36, 37 & Adam & $2.538958\!\cdot\!10^{-3}$ \\
        \hline
        Cascade NN & 26, 36, 48 & Adam & $1.651993\!\cdot\!10^{-3}$ \\  
        \hline
    \end{tabular}
\end{center}
\end{table}
As can be seen in Table~\ref{tab:optimResults}, the results obtained by optimising the hyperparameters are very close to those obtained without optimisation.
\begin{table}[t]
\caption{\label{tab:optimResults}Loss in 10 epochs obtained using optimised hyperparameters}
\begin{center}
    \begin{tabular}{|c||c|}
    \hline
        \multicolumn{1}{| c ||}{} &
        \multicolumn{1}{ c|}{Full test loss (MSE)} \\
        \hline \hline

        Single NN & $2.376151\!\cdot\!10^{-4}$ \\ \hline 
        Multiple NNs & $2.418988\!\cdot\!10^{-4}$ \\
        \hline 
        Cascade NN & $2.111828\!\cdot\!10^{-4}$\\

        \hline
    \end{tabular}
\end{center}

\end{table} 
This means that the parameters initially chosen were already good enough. Furthermore, better optimisation would require larger search spaces and more trials, which would mean much longer run times. However, we need to consider the computational cost and make a trade-off between it and the accuracy of the prediction.
Instead, by optimising the hyperparameters of the NN cascade, a slightly significant improvement and a lower test loss are obtained, confirming the validity of this model.

\subsection{Discussion}

Despite the fact that the error in terms of MSE is always of the order of $10^{-4}$, the results obtained allow us to make some considerations and comparisons between the neural structures tested and the corresponding optimised networks.
First of all, it should be noted that the pre-processing phase leads in any case to a significant performance improvement, up to a reduction of the test loss by $1.0\!\cdot\!10^{-4}$. Regarding the first experimental phase, in which the hyperparameters were chosen by trial and error, the best results were unexpectedly obtained by using a single NN including all the input and output data in the same model. This is an unexpected result, since it would be natural to think that the other two types of structure, which exploit the dependencies between the joint torques, would be more accurate. However, a possible explanation for this phenomenon is that although the study of the dataset shows significant independencies, the joints are never completely independent of each other.
That is why even the cascade NN performs better than the three separate networks. On the other hand, the parallel training of less complex networks involves a shorter run time, which is another aspect to take into account when dealing with large datasets. 

In the optimisation phase, after various tests, the hyperparameters that minimise the test loss are identified for each network. In general, there is not much improvement over the previous case, which means that the hyperparameters selected by trial and error were already good enough.
Instead, the effectiveness of exploiting the independent groups of joints is highlighted here. Indeed, after the optimisation process, the model that gives the most accurate results is the cascade network.
This demonstrates the correct functioning of a model that exploits the dependencies between joints, while keeping a link between the three groups. Again, although the difference in error is not very significant, the single NN gives slightly better results than the multiple networks, indicating that the influence of one group of joints on the others is probably not completely negligible. Another useful consideration is that, due to time and computational constraints, the number of trials and the range of hyperparameter values were limited in the optimisation process. In addition, the motion profiles were run with a limited range of velocities and accelerations, so better performance is expected within this range. 
This is not a problem if the required application is repetitive and has a limited range of joint positions, velocities and accelerations.

%% file: Conclusions.tex
\section{CONCLUSION AND FUTURE WORK}

This paper presents an effective approach to predict the joint torques of a robotic arm without relying on traditional model-based methods, which often require simplifications and a precise knowledge of all the system parameters. By using black-box models trained directly from experimental data - specifically NNs - the study provides promising results in accurately identifying nonlinear dynamics. The experiments were carried out with different types of NN architecture, including single NNs, multiple NNs, and cascade NNs. The results show that well-modelled NNs trained with appropriately pre-processed input data are able to accurately predict joint torques, even in the presence of parameter uncertainties and unconsidered dynamics such as friction and joint flexibility. 
%
The prior knowledge of the dependencies between joints not only reduces the computational cost, but also improves the prediction accuracy. Although hyperparameter optimisation yielded no significant improvements in error prediction due to already well chosen parameters, it validates the effectiveness of the cascade NN model in exploiting joint dependencies and linking all variables. 
%

Future work may focus on further refining NN architectures and exploring additional pre-processing and optimisation techniques to improve robustness and generalisability, and to develop more intelligent and adaptive robotic systems.
Moreover, despite their computational efficiency, NNs could lack physical plausibility. 
Physics Informed Learning (PIL) methods, integrating physical laws into black-box models, seem to be a promising way to improve their performance, but require further research~\cite{lutter2023combining}. Investigating PIL models for Lagrangian systems, also incorporating contact forces, are possible directions for improved efficiency, generalisation and physical plausibility. Strategies to reduce dataset size while keeping significant information may also be useful for improving dynamic model accuracy and productivity in robotics applications.